\pdfoutput=1

\documentclass[11pt]{article}

\usepackage[final]{acl}

\usepackage{times}
\usepackage{latexsym}
\usepackage[T1]{fontenc}

\usepackage[utf8]{inputenc}

\usepackage{microtype}

\usepackage{inconsolata}

\usepackage{graphicx}
\usepackage[font=small]{caption}
\usepackage[normalem]{ulem}
\useunder{\uline}{\ul}{}
\usepackage{algorithm}
\usepackage{algorithmic}
\usepackage{enumitem}
\usepackage{newfloat}
\usepackage{authblk}
\usepackage{listings}
\usepackage{tabulary}
\usepackage{amsmath}
\usepackage{booktabs}
\usepackage{multirow}
\usepackage{float}
\usepackage{indentfirst}
\usepackage[capitalize]{cleveref}
\usepackage{xcolor}
\usepackage{tcolorbox} 
\usepackage{hyperref}

\title{CogniBench: A Legal-inspired Framework and Dataset for Assessing Cognitive Faithfulness of Large Language Models}

\author[1]{\textbf{Xiaqiang Tang}}
\author[2]{\textbf{Jian Li}\thanks{Corresponding Authors}}
\author[1]{\textbf{Keyu Hu}}
\author[2]{\textbf{Du Nan}}
\author[2]{\textbf{Xiaolong Li}}
\author[3]{\\\textbf{Xi Zhang}}
\author[4]{\textbf{ Weigao Sun}}
\author[1]{\textbf{Sihong Xie}\textsuperscript{*}}
\affil[1]{The Hong Kong University of Science and Technology (Guangzhou)}
\affil[2]{Hunyuan AI Digital Human, Tencent}
\affil[3]{Beijing University of Posts and Telecommunications}
\affil[4]{Shanghai AI Laboratory}

\begin{document}
\maketitle
\begin{abstract}




Faithfulness hallucinations are claims generated by a Large Language Model (LLM) not supported by contexts provided to the LLM. Lacking assessment standards, existing benchmarks focus on ``factual statements'' that rephrase source materials while overlooking ``cognitive statements'' that involve making inferences from the given context.
Consequently, evaluating and detecting the hallucination of cognitive statements remains challenging.
Inspired by how evidence is assessed in the legal domain, we design a rigorous framework to assess different levels of faithfulness of cognitive statements and introduce the \textbf{CogniBench} dataset where we reveal insightful statistics. To keep pace with rapidly evolving LLMs, we further develop an automatic annotation pipeline that scales easily across different models. This results in a large-scale \textbf{CogniBench-L} dataset, which facilitates training accurate detectors for both factual and cognitive hallucinations.
We release our model and datasets at: \url{https://github.com/FUTUREEEEEE/CogniBench}


\end{abstract}

\section{Introduction}

With the widespread deployment of Retrieval Augmented Generation \cite{RAG,jokinen2024need}, Large Language Models (LLMs) \cite{chowdhery2023palm,openai2023gpt,touvron2023llama} are increasingly expected to generate responses that adhere closely to provided context. Consequently,   ``faithfulness hallucination'' \cite{survey_taxonomy,es2024ragas,saad2024ares} has become a critical research problem for modern LLMs applications.
 Prior work \cite{niu2023ragtruth,fava,belyi2024luna} attempted to address \textbf{factual} inconsistencies, which can be directly verified by comparing the model response with the provided context.

 \begin{figure}[ht!]
     \includegraphics[width=\linewidth]{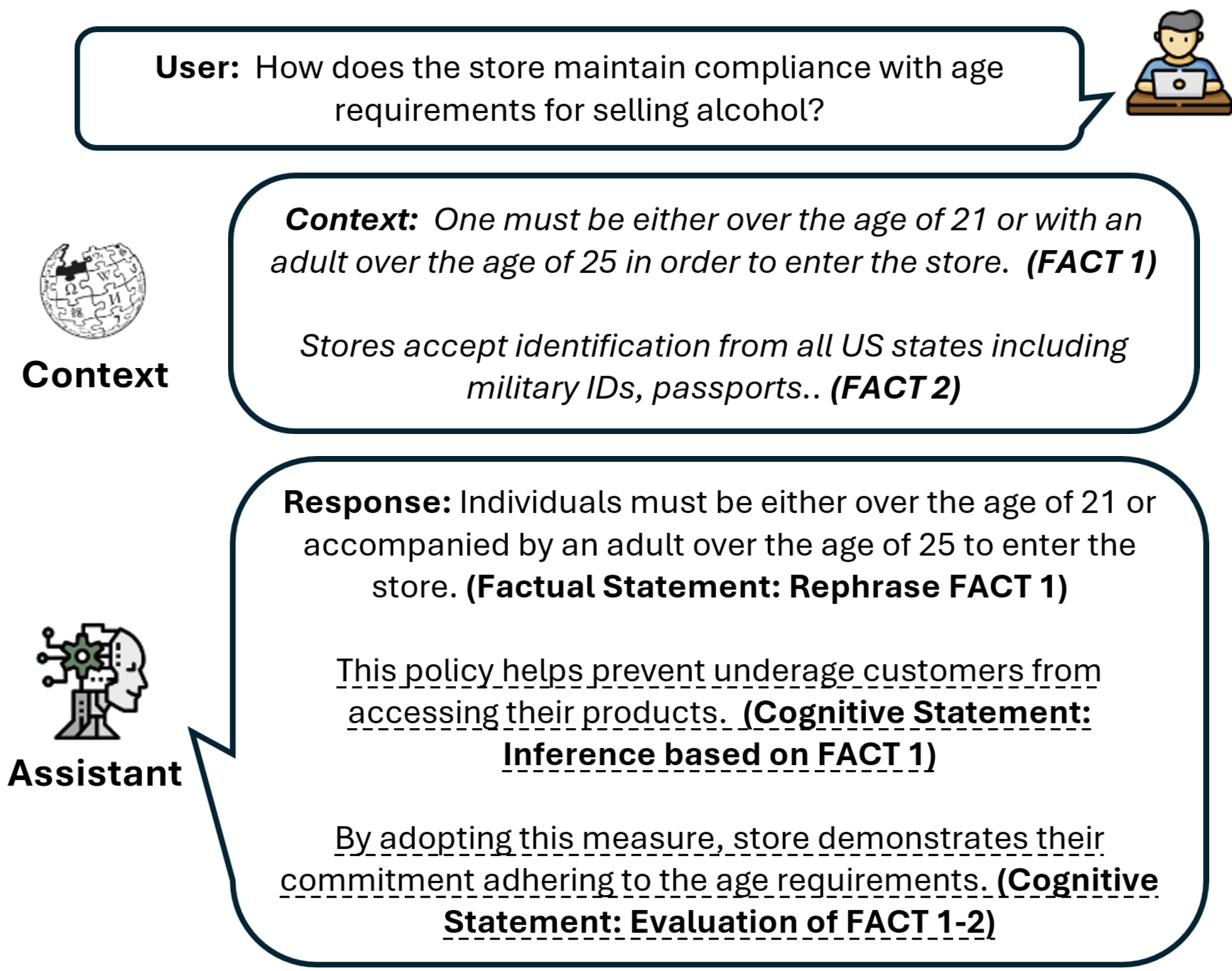}
     \caption{ Difference between ``factual statement'' and ``cognitive statement''. The former typically reproduces source materials, while the latter extends beyond the provided context to make inferences, provide explanations, or express evaluations and opinions. }
     \label{background}
\end{figure}

\begin{figure}[ht]
     \includegraphics[width=\linewidth]{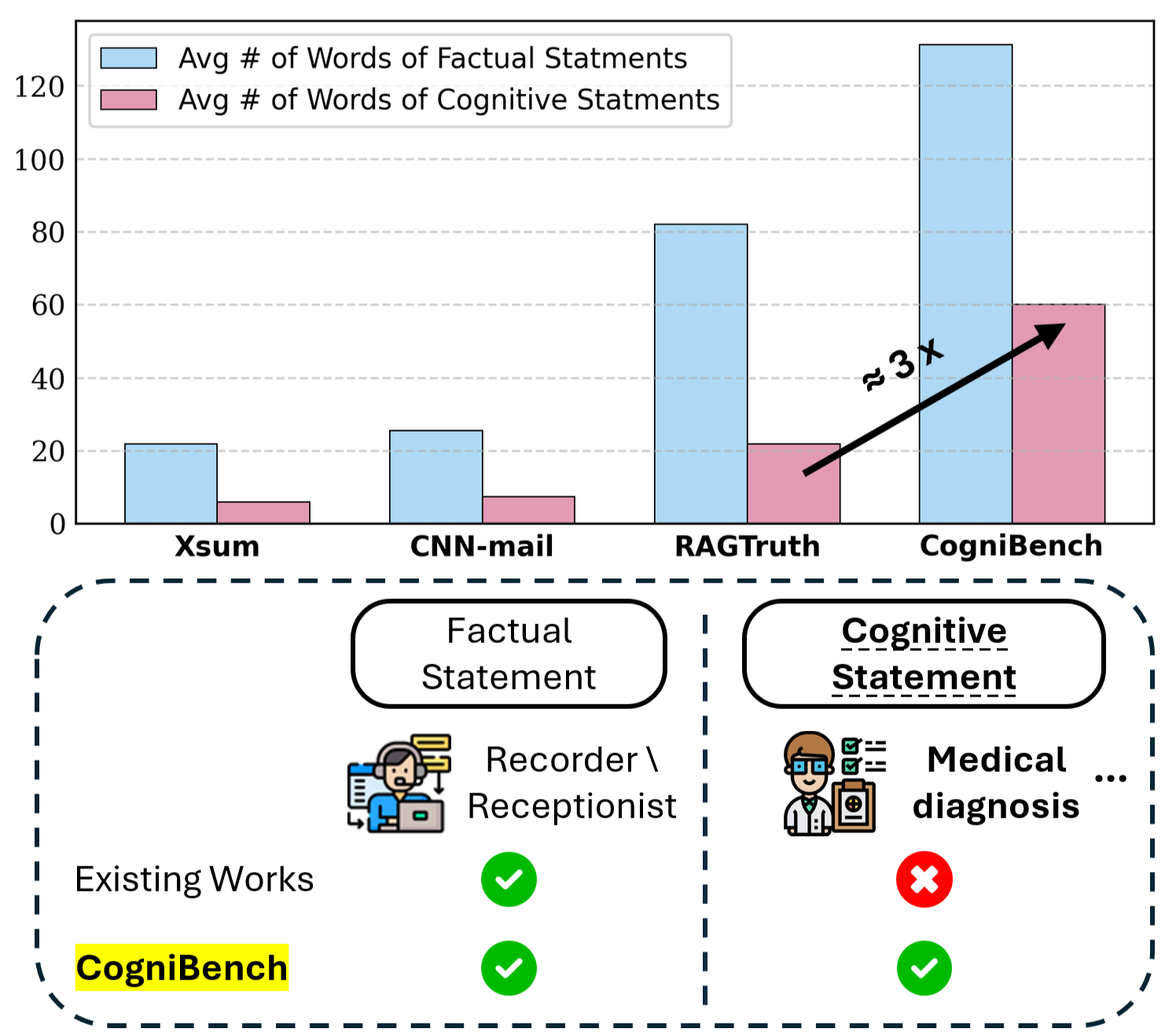}
     \caption{Dialogues in CogniBench on average contains three times more cognitive statements than previous datasets \cite{niu2023ragtruth,cnnmail,xsum}. While factual statements mirror human roles like recorders, cognitive statements require higher-level cognitive inference beyond simple rephrasing in applications such as medical diagnosis.}
     \label{gap}
\end{figure}

However, as LLMs evolve, LLMs increasingly generate \textbf{cognitive statements}— such as inferences, evaluations, and explanations that extend beyond the verbatim context \cite{webb2023emergent,zheng2023judging,xu2023large}. This is essential for increasing applications in medicine \cite{llm_doctor}, legislation \cite{llm_doctor}, and finance \cite{llm_lawyer}.  For instance, a medical AI must not only recall symptoms from a patient report but also draw diagnostic conclusions. Put in a broader perspective, we categorize LLM-generated statements based on Bloom’s Taxonomy \cite{bloom1956taxonomy}: \textbf{factual statements} (restate facts from context) and \textbf{cognitive statements} (applying, explaining, or evaluating knowledge), as exemplified in \cref{background}. 
Existing benchmarks \cite{fava,niu2023ragtruth,xsum}, which prioritize factual consistency, are primarily dominated by ``factual statement'' as shown in \cref{gap}.

Assessing cognitive statements generated by LLM becomes both important and challenging:
(C1) The assessment of cognitive statements lacks data and standardization. Their assessment is inherently subjective (e.g., determining whether an explanation is justified) and context-dependent—requirements can vary significantly across domains (e.g., creative storytelling \cite{llm_character} vs. clinical diagnosis \cite{llm_doctor}). 
(C2) Manual annotation is impractical for rapidly updated LLMs, necessitating low-resource, automated assessment methods applicable to both factual and cognitive statements.
We address these challenges through three key contributions:

\begin{itemize} [leftmargin=*]

\item In response to (C1), we introduce \textbf{CogniBench} (\cref{sec:Benchmark}), the first knowledge-grounded dialogue dataset and framework for assessing cognitive faithfulness. Built through rigorous manual annotation, CogniBench provides sentence-level annotations with a \textbf{legal-inspired} assessment protocol to reduce subjectivity. To accommodate diverse application requirements, we define three increasingly rigorous faithfulness criteria: \textit{rational} (plausible but unverifiable speculation), \textit{grounded} (contextually supported evaluations), and \textit{unequivocal} (undeniable conclusions). 
The increasing levels of rigorousness cater to the diverse needs of various LLMs' applications, from creative storytelling \cite{llm_character} to high-stakes medical diagnosis \cite{llm_doctor}.

\item In response to (C2), we develop an automated annotation pipeline (\cref{sec:auto_labeling}) that generates \textit{CogniBench-L}—a large-scale extension containing over 24k dialogues with sentence-level hallucination annotations. This resource supports the training of specialized hallucination detectors and enables the systematic benchmarking of evolving LLMs.

\item Empirical analysis reveals three pivotal insights: (1) cognitive statements increase with dialogue length (growing from 15\% in initial turns to 50\% by final turns \cref{hallu_rate_vs_len}), (2) LLMs exhibit \textbf{4.6× higher hallucination rates} when generating cognitive statements compared to factual ones (\cref{hall_rate_dis}), and (3) Model reliability varies drastically—while GPT-4 produces 33.7\% cognitive statements with a 60.1\% hallucination rate, Gemini-Pro generates 49.9\% cognitive content but suffers a 79.9\% error rate (\cref{llms_cog_pref}).

\end{itemize}

Furthermore, existing hallucination detectors struggle to assess cognitive statements accurately, with F1 dropping by 31\% compared to factual cases (\cref{eval_existing_hallu_det_model}). These findings emphasize the urgency of going beyond fact-centric assessments.

\begin{figure*}[ht]
     \includegraphics[width=\textwidth]{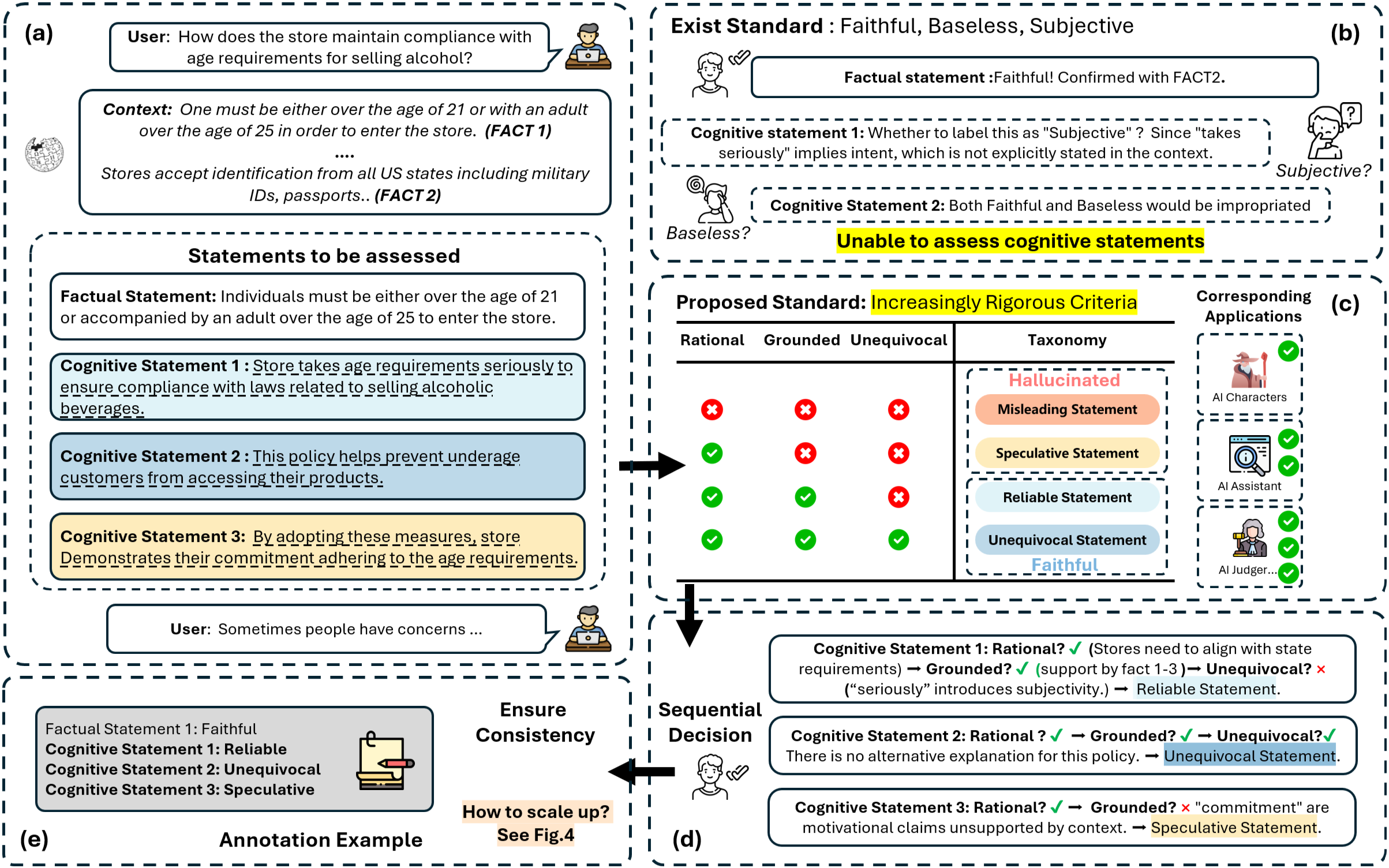}
     \caption{ (a)(b): Existing faithfulness assessment standards such as ``Baseless'' \cite{niu2023ragtruth} and ``Subjective'' \cite{fava} are ambiguous and insufficient for assessing cognitive statements. (c): We propose three increasingly rigorous assessment criteria (i.e. \textbf{Rational, Grounded, Unequivocal}) to annotate cognitive statements. (d)(e): Human annotators assess each statement in (a) by making sequential decisions. Cognitive statements can be categorized as Misleading, Speculative, Reliable, or Unequivocal based on what criteria are met. (See \cref{credibility:standards} for details)}
     \label{main}
\end{figure*}

\section{The CogniBench Benchmark}
\label{sec:Benchmark}

Knowledge-grounded conversations have become a crucial scenario in which people interact with LLMs. \cite{RAG,webgpt,llm_doctor}. 
Existing datasets primarily focus on assessing the faithfulness of ``factual statements'', typically assessing how well models rephrase information drawn from a given context. 
Moreover, the requirements for cognitive faithfulness—such as the ability to accurately make inferences, evaluations, or explanations—differ significantly across various application domains due to the diverse uses of LLMs. CogniBench aims to address these limitations by:
\begin{itemize}[leftmargin=*]
    \item Annotate a multi-round conversation dataset on the sentence level using the framework with detailed provenance information. The dataset can be used for understanding cognitive statements and detecting inconsistency (i.e. cognitive hallucinations) in LLMs. 

  \item Establishing a legal-inspired, increasingly rigorous assessment framework, which proposes three increasingly rigorous assessment criteria. This framework minimizes annotation efforts and reduces inconsistencies, offering users the flexibility to balance creativity and faithfulness according to their specific use cases.

\end{itemize}

\subsection{Overview of Dataset Creation}
\label{sec:data_creation}

 CogniBench leverages the approach from \cite{yang2023refgpt} to generate customized, multi-turn knowledge-grounded dialogues, minimizing factual errors. This enables us to focus on cognitive inconsistency. We use GPT-4 \cite{openai2023gpt} to generate a large corpus of dialogues, ensuring the conversations closely align with real-world LLM usage.
The guidelines for annotators to assess cognitive statements are summarized as follows:

\textbf{1. Identifying Statements Types:} First identify whether a statement rephrases facts from context (factual statement) or the statement extends beyond the provided context, such as inference, explanation, and opinions (cognitive statement).

\textbf{2. Assessing Faithfulness:} For factual statements, annotators assess whether the statement is ``faithful'' (consistent with the provided context) or ``invented'' (factual hallucination not supported by provided context) \cite{fava,niu2023ragtruth}. For cognitive statements, we propose a new assessment framework where no established standard exists, as detailed in \cref{credibility:standards}. Full instructions can be found in \cref{full_instruction}.

\subsection{A Legal-Inspired Tiered Framework for Assessing Cognitive Statements} 

We compare existing standards with our proposed framework in \cref{main}. This framework is characterized by two key features: 1) objectivity, ensured by pre-established laws, and 2) an increasingly rigorous design with progressively higher standards of faithfulness.

The annotations are done at the sentence level. Unlike span-level annotations of hallucinated entities or relationships of factual statements \cite{niu2023ragtruth,fava}. Cognitive statements and their cognitive nature should be assessed based on the semantics of an entire sentence.

Factual statements can be easily verified by comparing them with the provided context. For instance, Factual Statement 1 (``Individuals must be either over ...'') can be validated against FACT 1, as shown in \cref{main} (a).

In contrast, cognitive statements extend beyond the provided context and require reasoning. Their evaluation tends to be more subjective due to their inherent nature. For example, Cognitive Statement 2 (``This policy helps prevent...'')  extends FACT 2 from the provided context. Existing standards such as ``baseless'' or ``subjective'' fail to label such statements appropriately.

To build standardized assessment criteria that ensure consistent annotation, we draw inspiration from the legal domain. In legal contexts, evidence is categorized into \textit{Direct evidence} and \textit{Circumstantial evidence} (i.e. Indirect evidence).
As illustrated in \cref{equality}, \textit{Direct evidence} refers to when a witness directly describes the original event, whereas \textit{Circumstantial evidence} involves a witness suggesting a fact through indirect inference rather than direct observation.
 Similarly, cognitive statements in reasoning processes often infer conclusions from the provided context. Both require thorough analysis within their respective context to avoid misinterpretation. 
\textbf{We argue that the legal framework for validating \textit{Circumstantial evidence} offers a natural analogy for assessing the faithfulness of cognitive statements.} We further elaborate on these comparisons in \cref{append:equlity}.

Inspired by how the law validates \textit{Circumstantial evidence}, we propose three criteria (\cref{main} (c)) to assess the faithfulness of cognitive statements.

\phantomsection
\label{credibility:standards}

    \textbf{Criterion 1 Rational:} Whether a statement is reasonably believed, even if it does not have direct evidence from the provided context. 
    This criterion is derived from the legal distinction between \textit{inference} with \textit{speculation}, as outlined in \cref{append:inference_vs_speculation}. Specifically, \textit{``An inference that does not properly flow from the established fact is mere speculation.'' }
    Virtual AI characters \cite{llm_character} should adhere to this standard. Even though their primary goal may be entertainment rather than accuracy, their responses should not mislead users.

    \textbf{Criterion 2 Grounded:} Whether a statement can be supported by the provided context. If a statement satisfies criterion 1, the annotator will proceed to assess whether the statement can be logically derived from the available information.
    This is established from \textit{``An inference is a deduction of fact that may logically and reasonably be drawn from another fact or group of facts found or otherwise established in the proceedings.''}. 
    AI assistants like WebGPT \cite{webgpt} should meet at least this standard to ensure their responses are both helpful and grounded by context.
    
    \textbf{Criterion 3. Unequivocal:} 
    Whether a statement has no alternative interpretations could reasonably exist, free from subjective bias.
    This criterion is based on the legal requirement that \textit{  ``circumstantial evidence may be introduced, but the trier-of-fact must be satisfied beyond a reasonable doubt that the guilt of the accused is the only reasonable inference from the proven facts''} \cite{only_reasonable}.
    Responses from AI systems such as AI Judger, AI Doctor, or AI Trader \cite{llm_lawyer,llm_doctor,llm_financial_agent} should meet this standard. These systems make high-stakes decisions that directly impact human lives and livelihoods. Therefore, LLMs must make unequivocal statements to minimize the risk of error or bias.

With these three criteria, a statement can be categorized into one of the following: \textbf{Misleading}, \textbf{Speculative}, \textbf{Reliable}, or \textbf{Unequivocal}. 
For example, Cognitive Statement 1, labeled as Reliable (which might have been categorized as subjective or a hallucination under previous standards) is both Rational and Grounded, yet contains subjectivity. Despite this, it remains useful for AI assistant applications.
Additional annotated examples are presented in \cref{statment_rules}.

\begin{table}[h]
\resizebox{\columnwidth}{!}{
\begin{tabular}{ccc}
\hline
Annotation Method                                                                 & IAA & QA Instances                \\ \hline
\begin{tabular}[c]{@{}c@{}}Independent Multi-\\ Class Classification\end{tabular} & 91.51\%        & 25 (15 real-time QA + 10 post-hoc feedback) \\
\begin{tabular}[c]{@{}c@{}}Sequential Decision  \\ Framework\end{tabular}      & 96.19\%        & 13 (6 real-time QA + 7 post-hoc feedback )  \\ \hline
\end{tabular}}
\caption{Inter-Annotator Agreement (IAA) and Quality Assurance (QA) Effort (per 500 statements): Our Sequential Decision Framework achieves higher degree of agreement with less annotation effort.}
\label{hierarchical_vs_independent}
\end{table}

Our framework organizes the annotation process into a sequential decision-making structure (see full annotation protocol in \cref{annotation_protocol}), guiding annotators through a sequential decision-making process to classify statements. The design ensures the annotator only considers a more rigorous criterion (e.g., grounded or unequivocal) once previous less rigorous conditions (e.g., rational) are met, reducing cognitive load and the potential for conflicting interpretations. As shown in \cref{hierarchical_vs_independent}, this approach outperforms independent classification (i.e., direct ask annotator to categorize a statement into one type such as misleading, reliable) , achieving a 96.19\% inter-annotator agreement (IAA) while also reducing QA instances by 48\%. 

\begin{figure*}[ht]
     \includegraphics[width=\textwidth]{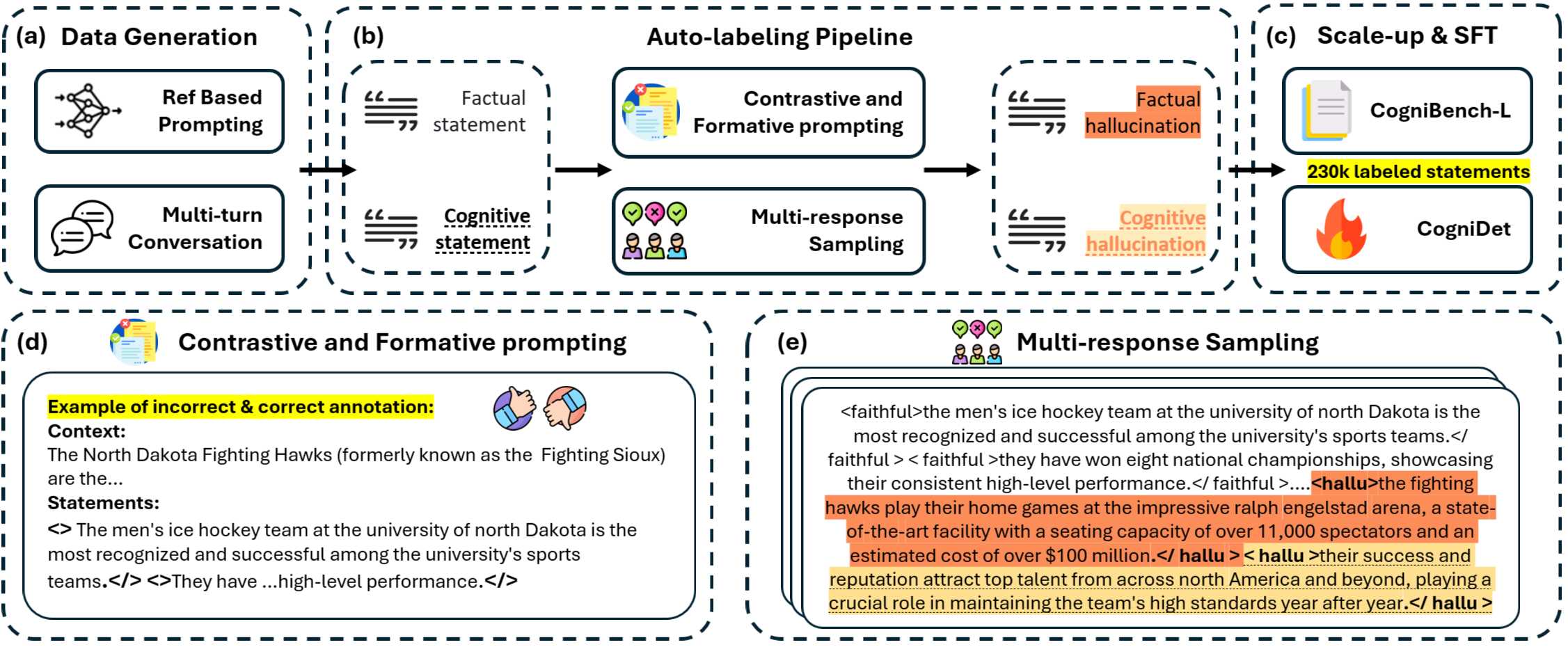}
     \caption{Auto-Labeling pipeline for hallucination detection. (a): We generate multi-turn, knowledge-grounded conversations using \cite{yang2023refgpt}. (b): A trained model classifies statements as factual or cognitive. Two prompting strategies are employed to generate sentence-level hallucination annotations with GPT-4. (c): This process results in a large-scale hallucination dataset CogniBench-L. We fine-tune an detection model to identify hallucinations in both factual and cognitive statements.}
     \label{method}
\end{figure*}

\section{Automated Data Expansion}
\label{sec:auto_labeling}

The rapid emergence of new LLMs poses significant challenges for faithfulness assessment, as manual annotation is time-consuming and costly. Moreover, relying on small-scale datasets is insufficient for the fine-tuning a generalized hallucination detection model \cite{niu2023ragtruth}. While synthetic hallucination texts \cite{fava,halludial} can guide LLMs to generate a considerable amount of annotated hallucinated data, they create a gap between simulated and real-world scenario.

To address this, we propose a fully automated labeling method that utilizes LLMs as judges to label dialogue data generated by advanced LLMs. This cost-effective approach can assess the faithfulness of new models and provides an scalable way to expand the CogniBench dataset into CogniBench-L, a comprehensive training corpus for hallucination detection model.

Results presented in \cref{Tab:ablation} show that Auto-Labeling can matches human annotators' accuracy making it a reliable proxy for assessing new LLMs. Additionally, Using the large-scale CogniBench-L dataset, we fine-tuned an 8B model CogniDet, which achieves state-of-the-art performance in detecting hallucinations in both factual and cognitive statements, as shown in \cref{eval_existing_hallu_det_model}. CogniDet is effective for low-cost hallucination detection in daily applications.

\subsection{Contrastive and Formative Prompting}
\label{CFP}

Achieving reliable sentence-level annotation of LLM outputs presents two primary challenges: (1) enabling the annotation LLM to accurately understand the standards outlined in \cref{credibility:standards}, and (2) the high cost of annotating sentences individually within dialogues.

Building on in-context learning \cite{LLM_few_shot_learner,survey_in_context_learning}, we provide the LLM with annotated examples that demonstrate both the desired task and the expected output format. Unlike conventional labeling tasks,  internalizing our faithfulness criteria is complicated by the inherent ambiguity of natural language. To address this, we employ a two-step process:

1. \textbf{Initial Prompting and Diagnosis}: We first prompt LLM with annotation examples from human annotators and ask the LLM to annotate statements accordingly. We then identify and collect common annotation errors.
For example, when labeling factual statements, the LLM may misclassify world-knowledge-verifiable facts not mentioned in the context as faithful.  When labeling cognitive statements, it may mistakenly classify logically deduced conclusions (e.g., Cognitive Statement 2 in \cref{main} (a) ``This policy helps prevent underage customers from accessing their products'') as hallucinations, even if they are contextually justifiable. 

2. \textbf{Contrastive Examples}: Based on these observations, we provide positive and negative examples in the prompt to clarify the boundaries between faithful statements and hallucinated statements.

 To avoid labeling sentences individually, we implement a formative prompting approach that guides the LLM to annotate every sentence in a single response. We use the NLTK toolkit \cite{nltk} to tokenize the text into sentences, then enclose each sentence with HTML-like markers `<>` and `</>`. The LLM is then instructed to place the assessment for each sentence within these markers (e.g., <faithful>, <hallu>). Prompts are shown in \cref{appendix:prompt}. For cognitive statements, a second-stage prompt further categorizes them into four sub-categories.

\subsection{Multi-response Sampling}
\label{Sampling}

Although Contrastive and Formative Prompting can significantly reduce the ambiguity in defining cognitive faithfulness, individual LLM responses may still manifest hallucinatory judgments. To further enhance reliability, we adopt a sampling-based technique inspired by recent hallucination detection methods \cite{semantic_entropy}.

In practice, we prompt the LLM five times for each annotated instance, maintaining identical prompts. This yields multiple independent assessments for the same sentence. We then apply a majority-vote criterion: a sentence is deemed faithful only if the majority of sampled responses classify it as such. By aggregating multiple judgments, this approach mitigates the impact of occasional hallucinations or outlier predictions, ensuring that the final annotated dataset more accurately reflects credible cognitive statements.

As shown in \cref{Tab:ablation}, our contrastive and formative prompting (CFP) and Multi-response Sampling (Sampling)  significantly enhance the LLM’s ability to discern faithful statements from hallucinations.

\section{Experiments}
\subsection{Matrices and Baseline}

    We report precision, recall, and the F1 score for detection results. For instance: 
$
{Recall} = \frac{{\#\ of\ Words\ in \ detected\ hallucination\ sentence}}{{\#\ of\ Words\ in\ labeled\ hallucinated\ sentence}}.
$ Precision and F1 scores are calculated using analogous formulas.
    Using CogniBench, we conducted experiments with the following six distinct algorithms for hallucination detection as detailed in \cref{app: baselines}.

\subsection{Statistic results of Cognibench}

\begin{table*}[h!]
\centering
\resizebox{\textwidth}{!}{
\begin{tabular}{@{}cccccc@{}}
\toprule
Dataset      & Num Response & Num Conversation & Num Labeled Sentences & Num Context Words (min-max (avg)) &  Words per Response \\ \midrule
CogniBench   & 264       & 179             & 2516                  & 297–1252 (696.94)                 & 50-432(200.44)               \\
CogniBench-L & 24084     & 7058            & 234164                & 8-1409 (711.71)                    & 8-709(201.38)                \\ \bottomrule
\end{tabular}}
\caption{The basic statistics of CogniBench.}
\label{basic_statistic}
\end{table*}

We present basic statistics of the CogniBench dataset in \cref{basic_statistic}. CogniBench is a human-annotated following the framework proposed in \cref{sec:data_creation}. It is the first benchmark specifically designed to assess the faithfulness of cognitive statements while also incorporating factual statement annotations. Unlike previous datasets \cite{niu2023ragtruth,fava}, CogniBench uniquely features multi-turn conversations with extensive and diverse contexts, enabling a comprehensive assessment of language models' cognitive reasoning.

Due to the scalability limitations of human annotation, we employ GPT-4 to generate CogniBench-L, an automatically labeled dataset, using the methodology outlined in \cref{sec:auto_labeling}. CogniBench-L is 100 times larger than  CongiBench.
This expansion facilitates the development of a more generalized and robust hallucination detection model. As demonstrated in \cref{Tab:CogDet}, models trained on CogniBench-L outperform existing faithfulness-based hallucination detection approaches, underscoring the effectiveness of our dataset in enabling improved assessment and mitigation of cognitive hallucinations in large language models.

\subsection{Cognitive Dynamics in Conversation}
\label{hallu_rate_vs_len}

\begin{figure}[ht]
     \includegraphics[width=\linewidth]{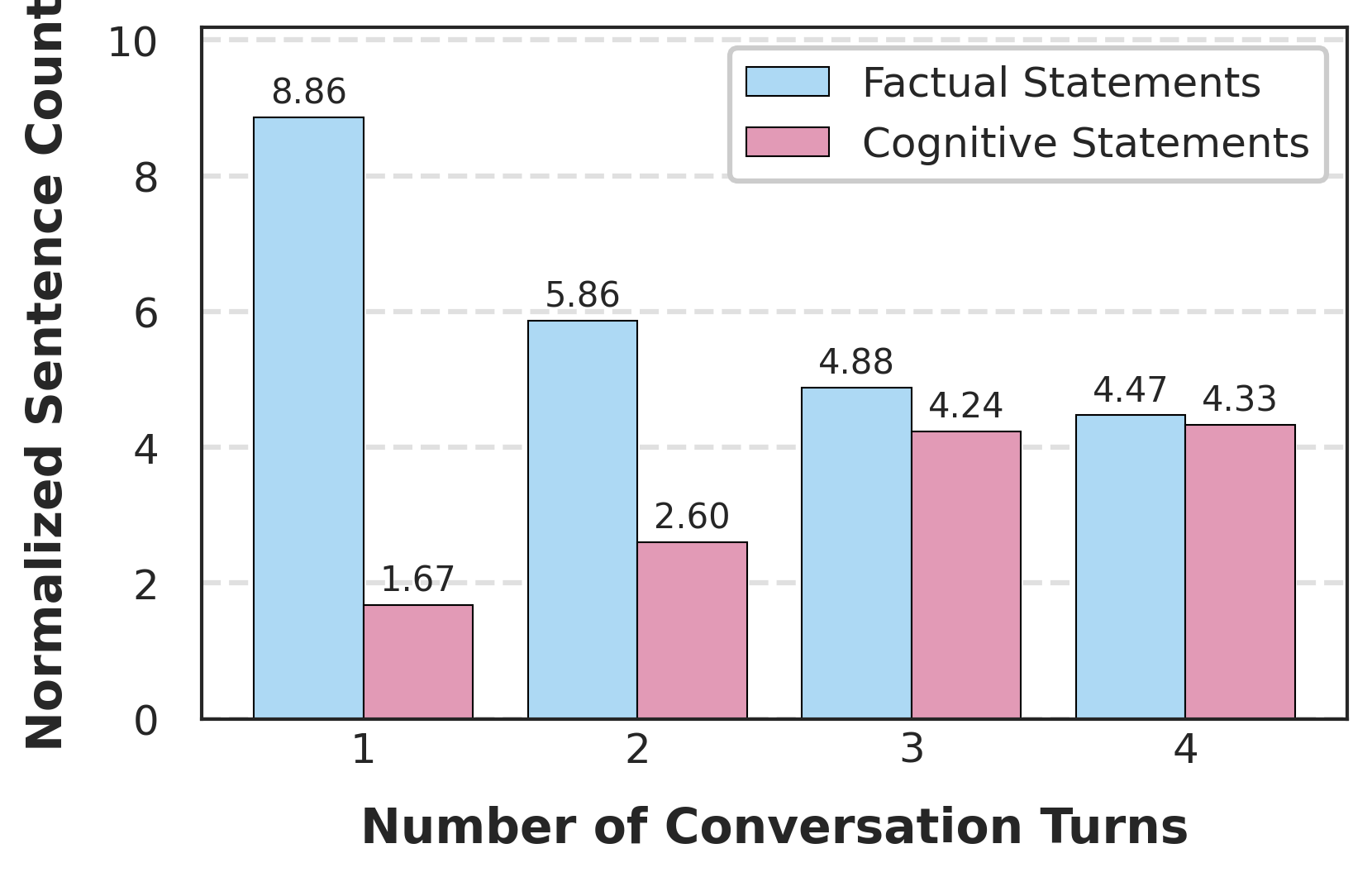}
     \caption{Average number of factual statements decreases as the number of conversation turns increases, while the number of cognitive statements increases with more turns.}
     \label{cog_dis_vs_length}
\end{figure}


\textbf{Cognitive statements increase with the length of conversation} In \cref{cog_dis_vs_length}, we illustrate the dynamics of factual and cognitive statements relative to the number of conversation turns.

The distribution of factual and cognitive statements across conversation turns reveals distinct patterns. LLMs initially follow the provided context closely and generate factual statements, but as the dialogue progresses, they produce more cognitive statements. Showing a shift toward deeper, more reflective interactions as the conversation unfolds.

This pattern underscores the significance of examining multi-turn, knowledge-grounded conversations, where the role of cognitive statements becomes more pronounced as the exchange develops. Current datasets, however, are either limited to single-turn conversations \cite{niu2023ragtruth} or are not long enough to capture cognitive statements effectively \cite{halludial}.


\subsection{Statement Distribution}
\label{hall_rate_dis}

\begin{figure}[h]
    \centering
     \includegraphics[width=0.85\linewidth]{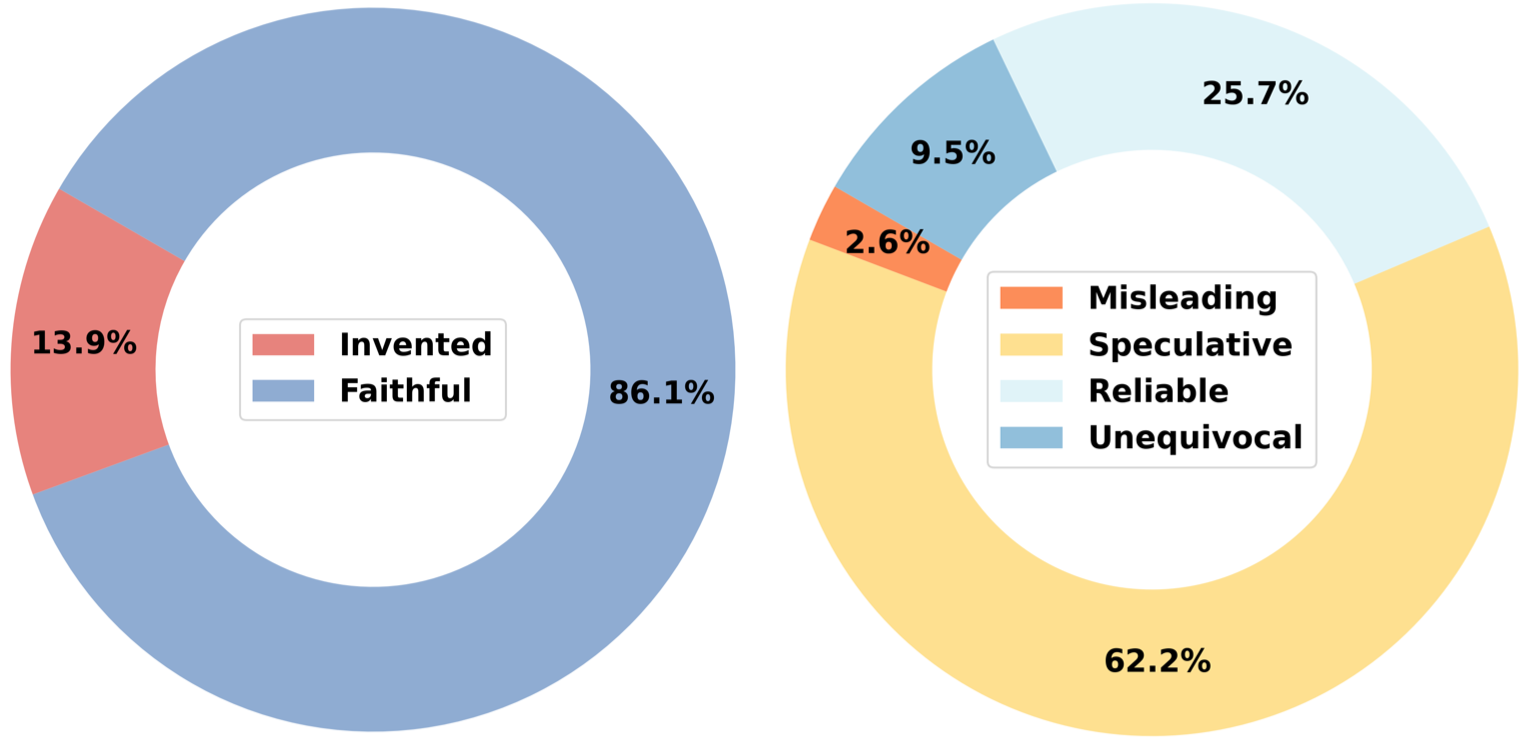}
     \caption{\textbf{Left:} Distribution of Factual statements. \textbf{Right:} Distribution of Cognitive statements. \textbf{LLMs are more likely to generate cognitive hallucinations (i.e., misleading and speculative statements) when producing cognitive statements.}}
     \label{dis_by_hallu_type}
\end{figure}


\begin{figure*}[h!]
     \includegraphics[width=\textwidth]{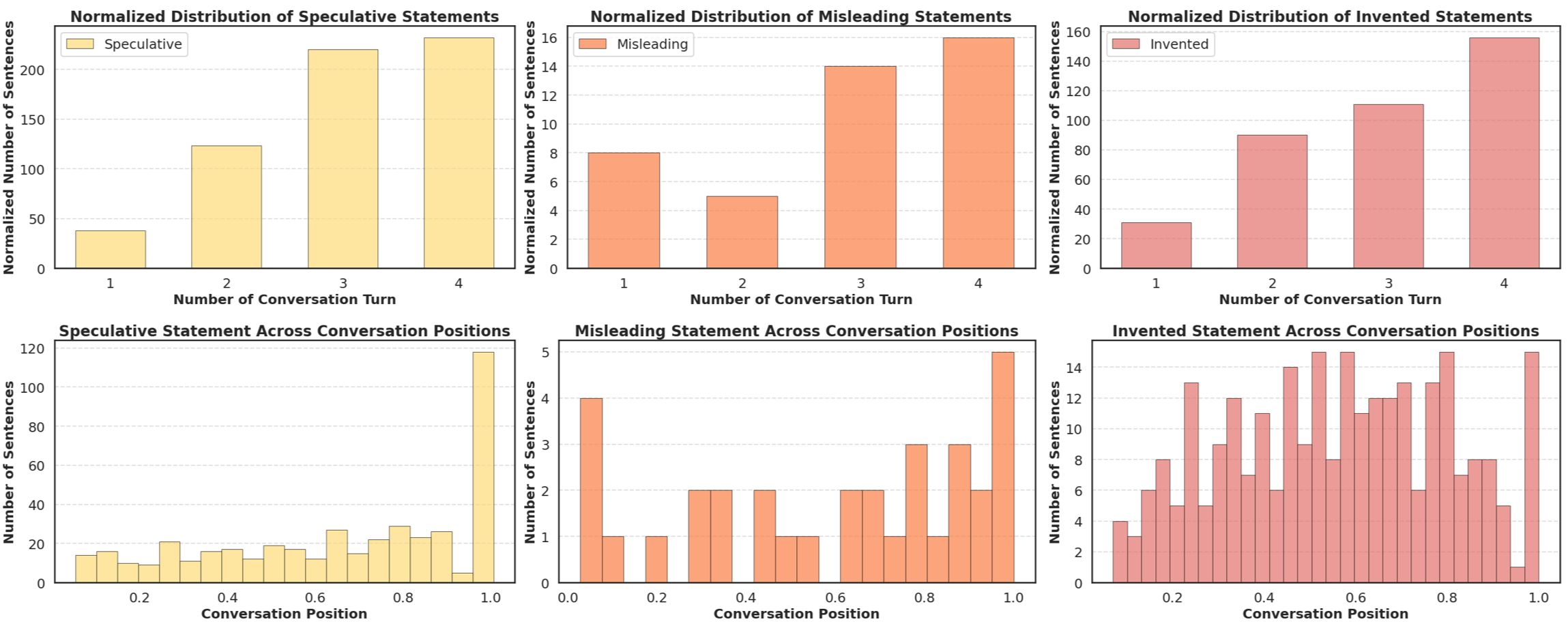}
     \caption{\textbf{Top:} Hallucinated statements increase with more conversation turns. \textbf{Bottom:} Distribution of hallucinated statements across dialogues. \textbf{Cognitive hallucinations (speculative or misleading statements) tend to occur at the beginning or end of a dialogue, while factual hallucinations are more likely to appear in the middle of the dialogue.}}
     \label{hallu_dis_pos}
\end{figure*}


\textbf{LLMs exhibit a higher risk of hallucination when generating cognitive statements compared to factual ones.}
We analyze the distribution of statements in CogniBench according to their faithfulness. For factual statements, 13.9\% were found to have no supporting sources from the context and were thus labeled as  ``invented'' by our annotators. In contrast, for cognitive statements, 62.2\% were categorized as speculative, satisfying only Criterion 1 in \cref{credibility:standards}. Furthermore, 2.6\% of cognitive statements did not meet Criterion 1 and were categorized as misleading.  Consequently, the overall hallucination rate for cognitive statements is 64.8\%, compared to 13.9\% for factual statements.

These findings highlight a key limitation of LLMs: while they are generally accurate in recalling and understanding factual content, they often struggle to apply this knowledge in cognitively demanding tasks, underscoring the urgent need to improve their faithfulness in such contexts.

\subsection{Faithfulness Dynamics in Conversation}

In \cref{hallu_dis_pos}, we examine the faithfulness hallucination occurrence positions. As shown in the top plot, both factual and cognitive hallucinations are significantly more likely to occur as the number of conversation turns increases. This observation is consistent with recent studies \cite{niu2023ragtruth, lost_in_middle, wang2024ada}, which suggest that the faithfulness of LLMs tends to decrease as the context lengthens. CogniBench further demonstrates that, as conversations progress, LLMs exhibit a consistent decline in faithfulness.

Additionally, the bottom plot presents the distribution of hallucinations within individual conversation turns. We observe that factual hallucinations tend to occur in the middle of a turn. In contrast, cognitive hallucinations are more likely to occur at the end of a turn, with some also happening at the beginning.
This pattern suggests that LLMs are more prone to introducing questionable conclusions or opinions at the beginning or end of a turn where hallucinations are especially likely to occur.

\subsection{Evaluate Hallucination Detection Methods on Cognibench}
\label{eval_existing_hallu_det_model}

\begin{table}[h!]
\resizebox{\columnwidth}{!}{
\begin{tabular}{@{}ccccc@{}}
\toprule
\multicolumn{2}{c}{\multirow{2}{*}{Method}} &
  \multirow{2}{*}{Overall} &
  \multirow{2}{*}{\begin{tabular}[c]{@{}c@{}}Factual \\Hallucination\end{tabular}} &
  \multirow{2}{*}{\begin{tabular}[c]{@{}c@{}}Cognitive \\Hallucination\end{tabular}} \\
\multicolumn{2}{c}{} & & & \\ \midrule
\multirow{2}{*}{Prompting} 
  & ChatGPT‑3.5  & 48.54 & 22.98 & 56.57 \\
  & ChatGPT‑4   & 58.03 & 46.82 & 66.04 \\ \midrule
\multirow{2}{*}{NLI}  
  & Tasksource (COLING 2024)      & 26.87 & 27.10 & 26.75 \\
  & SelfCheckGPT (EMNLP 2023)     & 45.81 & 32.08 & 61.10 \\ \midrule
\multirow{2}{*}{E2E}  
  & Fava (CoLM 2024)              & 7.90  & 12.90 & 5.10  \\
  & RAGTruth (ACL 2024)           & 23.90 & 45.30 & 11.20 \\ \midrule
\multirow{2}{*}{Ours} 
  & Auto-Labeling                & \textbf{82.20} & \textbf{82.50} & \textbf{81.90} \\
  & CogniDet 8B                  & \underline{70.30} & \underline{64.40} & \underline{73.80} \\ \bottomrule
\end{tabular}}
\caption{Hallucination detection performance (sentence‑level F1) on CogniBench. Cognitive hallucination refers to misleading and speculative statements. Overall scores represent macro‑averages across both categories.}
\label{Tab:CogDet}
\end{table}

We evaluate existing hallucination detection methods on CogniBench, revealing significant performance gaps between factual and cognitive hallucination detection. While Fava and RAGTruth achieve 12.9\% and 45.3\% F1 on factual hallucinations, respectively, their performance drops to 5.1\% and 11.2\% for cognitive hallucinations, highlighting CogniBench's challenging nature.

Auto-Labeling pipeline (\cref{sec:auto_labeling}) achieve performance with 82.2\% overall F1, closely matching human annotation and demonstrating reliable annotation capability. We fine-tune a 8B-parameter model CogniDet on the auto-labeled CogniBench-L corpus which reaches 70.3\% F1 with a single forward pass—far cheaper than NLI baselines that perform pairwise sentence–context comparisons. Complete detection examples are provided in \cref{appendix:CogniDet_example}. We believe auto-Labeling pipeline can serve as a proxy to assess new LLMs, and CogniDet is suitable for low-cost hallucination detection in everyday applications.


\subsection{Evaluate the Faithfulness of LLMs}
\label{llms_cog_pref}

\begin{figure}[h!]
     \includegraphics[width=\linewidth]{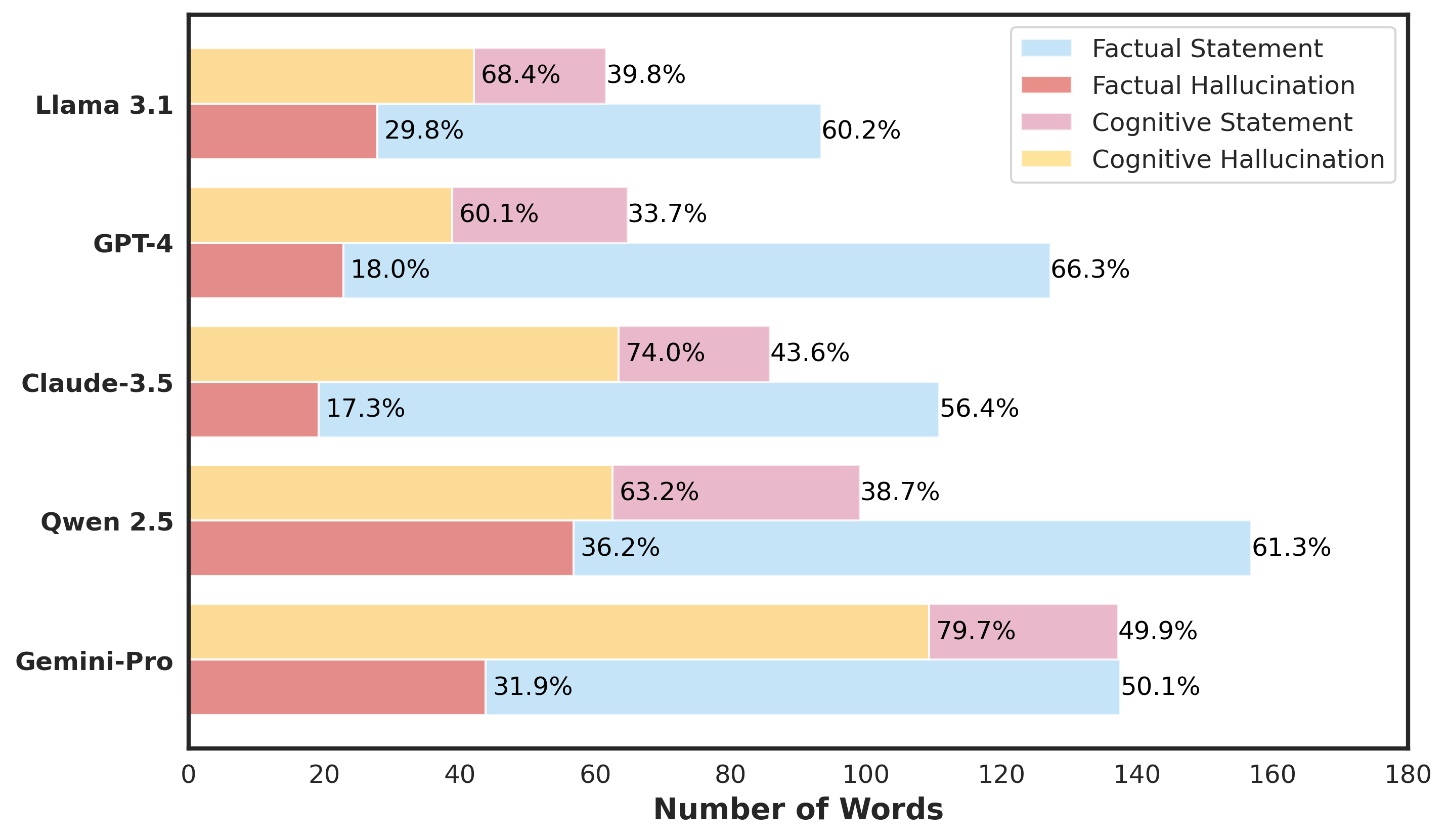}
     \caption{Evaluation of factual and cognitive statement portions with respect to their faithfulness via the proposed auto-labeling pipeline. Models evaluated: Llama 3.1 (Llama 3.1 70B), GPT-4 (GPT-4-1106-preview), Claude 3.5 (Claude-3-5-sonnet-20241022), Qwen 2.5 (Qwen2.5-72B-Instruct), and Gemini-Pro (Gemini-1.0-Pro).}
     \label{eval_exist_llm}
\end{figure}


\textbf{LLMs show distinct preferences for generating cognitive versus factual statements, and the reliability of their cognitive statements varies.} To evaluate this, we applied the proposed auto-labeling method to five popular advanced models, both closed-source and open-source, as shown in \cref{eval_exist_llm}. All models were prompted with the same input and followed the approach in \cite{yang2023refgpt} for generating knowledge-grounded conversations. 

Results show that GPT-4 generates the largest portion of factual statements, with 66.3\% of its response categorized as factual. On the other hand, Gemini-Pro tends to steer the conversation toward generating more cognitive statements, accounting for 49.9\% of its response. In terms of faithfulness, GPT-4 exhibits the highest reliability for cognitive statements but still suffers from hallucinations, with 60.1\% of cognitive statements identified as hallucinations. Claude-3.5 stands out for its high faithfulness in factual statements, with only 17.3\% of those statements being hallucinations.

\section{Conclusion}

In this paper, we introduced a legal-inspired framework and \textbf{CogniBench}, a novel dataset for assessing the faithfulness of LLMs in generating factual and cognitive statements. We establish an increasing level of rigorousness evaluation framework, catering to application-specific requirements.
We have analyzed of LLMs' dialogue patterns and the dynamics of their faithfulness. We found that while LLMs are generally capable of rephrasing factual information accurately, their reliability drops significantly when generating cognitive statements. This highlights the need to improve cognitive capabilities in LLMs.

Additionally, we propose an automated labeling method using LLMs as judges, facilitating access to a broad range of popular models. This method allows for the scalable expansion of \textbf{CogniBench-L}. We fine-tuned \textbf{CogniDet} on CogniBench-L, which achieved state-of-the-art performance in detecting hallucination in both factual and cognitive statements. 
We believe this work lays the groundwork for the future development of LLMs' cognitive capabilities, enabling more intelligent and safer   deployment in applications.

\section*{Limitations}

The study of hallucination in large language models (LLMs) is rapidly evolving. Our work presents a comprehensive framework for assessing the faithfulness of both factual and cognitive statements in LLMs. However,  several caveats remain.


\paragraph{Domain specificity.}
Our benchmark centres on common-sense, open-domain dialogues.  High-stakes applications (e.g., medical, financial, national-security) demand annotators with domain expertise and finer-grained taxonomies of error.  Extending our protocol to such domains will therefore require specialised guidelines and expert-curated test sets.

\paragraph{Legal analogy.}
Legal concepts vary across jurisdictions, jurisprudence, and judicial discretion.  We employ legal framework to structure cognitive-faithfulness judgements, not to impose rigid, jurisdiction-specific standards.  The analogy should be read as inspiration for systematic reasoning, not as a normative legal framework.

\paragraph{Source coverage and bias.}
Wikipedia offers broad coverage and a transparent revision history, making it a convenient starting point; nevertheless, its editorial biases may skew reference distributions.  Because our toolkit is fully open-source and modular, future iterations can plug in alternative corpora to mitigate such biases.


\section*{Ethical considerations}
This work is in full compliance with the Ethics Policy of the ACL. We acknowledge that responses generated by LLMs in this study may contain inaccuracies. Aside from this, to the best of our knowledge, there are no additional ethical issues associated with this paper.

\section*{Acknowledgments}
Sihong Xie was supported by the Department of Science and Technology of Guangdong Province (Grant No. 2023CX10X079), the National Key R\&D Program of China (Grant No. 2023YFF0725001), the Guangzhou-HKUST(GZ) Joint Funding Program (Grant No. 2023A03J0008), and Education Bureau Guangzhou Municipality.
\bibliography{custom}

\appendix

\section{Appendix}
\label{sec:appendix}

\subsection{Related Work}

\subsection{Hallucination of Large Language Models}

Hallucination in LLMs can be generally categorized into two types: factuality hallucination, where
generated content deviates from established world knowledge (e.g., claiming ``Mars has oceans''), and faithfulness hallucination, where the generated response is inconsistent with the provided context(e.g., misrepresenting a source document’s information). \cite{survey_taxonomy}

Existing research has demonstrated that incorporating up-to-date, relevant knowledge in the prompt can effectively reduce factuality hallucination \cite{RAG}.
In contrast, faithfulness hallucination persistent between the model’s response and the provided context. 

Early work focused on summarization tasks \cite{cnnmail,xsum}, but the rise of RAG systems \cite{RAG} has shifted attention to faithfulness in knowledge-grounded generation, where hallucination risks compound with contextual complexity.

\subsubsection{Faithfulness Hallucination Datasets}

Recently, \cite{fava} proposed a fine-grained taxonomy for faithfulness hallucinations in long-form text generation by synthesizing hallucinated examples. \cite{niu2023ragtruth} developed a word-level hallucination detection method specifically for RAG applications, leveraging manually annotated data.
However, existing datasets prioritize "factual statements," focusing on surface-level errors such as entity mismatches or paraphrasing inaccuracies.
Moreover, they are typically single-turn dialogues \cite{niu2023ragtruth} or short contexts dominate \cite{halludial}, neglecting multi-turn interactions where cognitive statements emerge (\cref{hallu_rate_vs_len}).

CogniBench addresses these gaps by curating dialogues rich in cognitive statements—claims requiring inference ("This policy prevents underage access"), evaluation, or hypothetical reasoning (Figure \ref{background}). These statements reflect real-world LLM applications in medicine, law, and finance, where unfaithful reasoning carries high stakes.

We argue that as LLMs continue to advance, the importance of "cognitive statement" will rise, and users will increasingly rely on their capabilities for complex reasoning and decision-making.

\subsubsection{Faithfulness Hallucination Detection}

Faithfulness assessment frameworks span tasks from summarization \cite{fabbri2022qafacteval,maynez2020faithfulness,scialom2021questeval} to knowledge-grounded dialogue \cite{niu2023ragtruth,fava,selfcheckgpt}.
However, these methods prioritize factual consistency (e.g., detecting entity/relation errors) and struggle with cognitive statements. For instance, an LLM might faithfully cite a medical report’s data (factual) yet draw an unfounded diagnostic conclusion (cognitive). Assessing the latter requires reasoning about contextual plausibility, not just textual overlap—a challenge existing tools are unequipped to address. 

Based on CogniBench, we developed a detection model that excels at detecting both hallucinations in "factual statements" and "cognitive statement" \cref{Tab:CogDet}.











\subsection{The Equality of Assessing Cognitive Statements and the validation of Circumstantial Evidence}
\label{append:equlity}

\begin{figure}[h!]
    \centering
     \includegraphics[width=\linewidth]{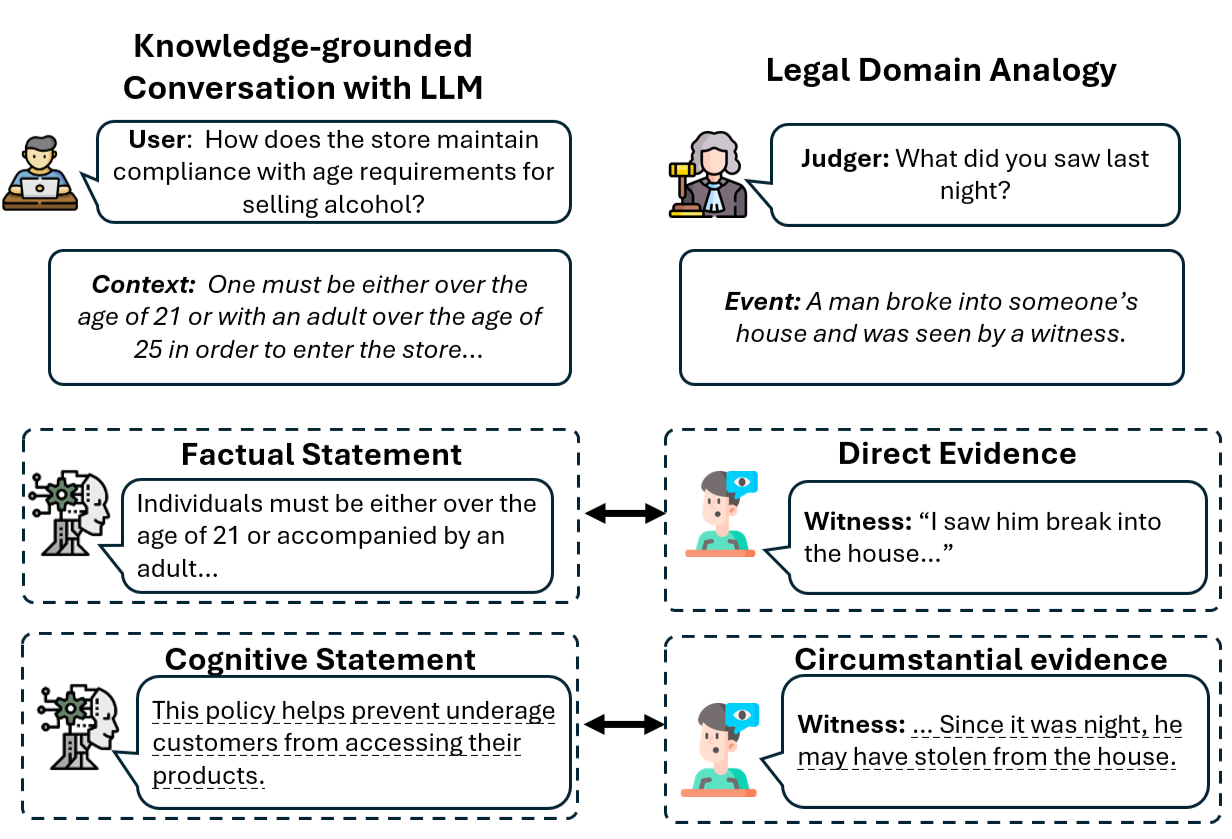}
     \caption{Drawing parallels between legal domain evidence and Cognitive Statements in knowledge-grounded conversations with the LLM}
     \label{equality}
\end{figure}

As shown in \cref{equality}, the process of evaluating cognitive statements in knowledge-grounded conversations can be viewed through the lens of legal domain evidence. In this analogy, the user assumes a role akin to a judge in a legal proceeding. The context provided by the conversation serves as the event that has occurred. 

Factual statements within a conversation directly align with \textit{Direct evidence} in a legal context. For example, when the user asks about age requirements in the store, a factual statement like "Individuals must be either over the age of 21 or accompanied by an adult" can be directly validated by comparing it with the context provided. This mirrors how direct evidence in a legal case directly points to the truth of the matter.

On the other hand, cognitive statements require more nuanced assessment. Just as \textit{Circumstantial evidence} suggests facts based on indirect observation, cognitive statements extend beyond the given context and rely on logical inferences. In the figure, cognitive statements like "This policy helps prevent underage customers from accessing their products" rely on reasoning drawn from the provided context, just as circumstantial evidence in a court case requires drawing reasonable inferences to support a conclusion.

By leveraging the long-established and reliable standards from the legal domain, we can ensure a more rigorous and systematic assessment of cognitive statements in knowledge-grounded conversations with LLMs.

\subsection{Definitions}
\label{append:definition}

We refer to \cite{garner2004black,nycourts_evidence,law2008criminal} for following definition:

(1) \textit{Direct evidence} is evidence of a fact based on a witness’s personal knowledge of that fact acquired by
means of the witness’s senses. \textit{Direct evidence} means evidence which immediately points to the question at issue.

(2) \textit{Circumstantial evidence }is indirect evidence that does not, on its face, prove a fact in issue but gives rise to a logical inference that the fact exists. \textit{Circumstantial evidence} requires drawing additional reasonable inferences in order to support the claim.

\phantomsection
\label{append:Drawing_inference}
(3)  \textit{Drawing inferences} can be described as a two-step process. The first step is to find that the facts from which the inference is to be drawn have been proven in the trial. If not then any inference is of necessity nothing more than speculation. The second step is to make an inference from the proven facts that is reasonable, rational and logical

\phantomsection
\label{append:inference_vs_speculation}
(4) \textit{Inference vs. Speculation:} 
An inferred fact must be one that is "reasonably and logically drawn from a fact or group of facts established by the evidence." An \textit{inference} that does not properly flow from the established fact is mere \textit{speculation}.

An \textit{inference} is a "deduction of fact that may logically and reasonably be drawn" from objective facts. \textit{Speculation} is when the judge theorizes without evidentiary support or where a conclusion is drawn in the absence of a proven fact.

\subsubsection{Examples of Direct and Circumstance Evidence}
\label{example_of_circums_evidence}

In a trial, if someone is trying to prove it rained on a certain morning, a witness can provide evidence in two ways.

\textit{Direct evidence}: If the witness says they walked to the subway, saw rain, felt it, and heard it, this is direct evidence that it rained that morning.

\textit{Circumstantial evidence}: If the witness says it was clear when they walked to the subway, but later saw people on the train with wet umbrellas and clothes, this doesn't directly prove it rained. However, it suggests, through logical inference, that it likely rained, making it circumstantial evidence.



\subsection{Details of CogniBench Annotation}
\subsubsection{Construction of Knowledge-grounded Dialogues}

\paragraph{Knowledge source.}
All background passages are retrieved exclusively from the 2025 Wikipedia \cite{wikipedia2025}.

\paragraph{Reference selection.}
Every candidate passage is first tagged with a topic label produced by \textsc{Llama-3-70B-Instruct}.  
We then draw an equal number of passages from each label to guarantee balanced topical coverage (see \cref{topic_distribution}).

\paragraph{Dialogue generation.}
We leverage the RefGPT approach~\cite{yang2023refgpt} to generate customized, multi-turn, knowledge-grounded dialogues while minimizing factual errors. User prompts are capped at \textasciitilde150 words, and assistant replies at \textasciitilde500 words—limits that preserve context-window budget yet allow fully referenced answers. Dialogue turns are authored by a mixture of state-of-the-art instruction-tuned models, with \textsc{GPT-4-1106-preview} serving as the primary engine alongside other large language models to ensure realistic and coherent interactions.

\subsubsection{Annotation Protocol}
\label{annotation_protocol}
Identifying hallucinations within cognitive statements is a challenging task that requires critical thinking and a deep understanding of the logical flow of text across various topics. To ensure the reliability of the annotation process, we have implemented the following measures:
\begin{itemize}[leftmargin=*]
    \item We hired annotators from a professional vendor. All annotators are proficient in English and possess at least a bachelor's degree.
     \item Annotators undergo comprehensive training to ensure they fully understand the annotation standards outlined in \cref{credibility:standards}. They must also pass a test annotation set before beginning formal annotation.
    \item Each sentence in CogniBench is reviewed by two annotators; in case of discrepancies, a third review is conducted.
    \item An online feedback and QA form is provided, allowing annotators to raise any questions or clarifications with us if they encounter any confusion during the annotation process.
\end{itemize}

Each annotator gets paid at a rate of 0.30\$ per statement assessed.

\subsubsection{Full Instruction for Annotator}
\label{full_instruction}

As shown in \cref{fig:full_instruct}, we provide comprehensive instructions for annotators to follow when labeling each dialogue. The instructions include steps for identifying irrelevant statements, classifying the type of statements, and assessing factual and cognitive statements. Additionally, we offer positive and negative examples to help annotators better understand the guidelines for each category. For simplicity, these examples are omitted here.

\begin{table*}[!ht]
\renewcommand{\arraystretch}{1.1}
\small
\centering
    \begin{tabular}{p{1.0\textwidth}}
    \toprule
    \textbf{\textsc{Step 1: Identify Irrelevant Statements}}\\
    \midrule
    A statement is irrelevant if it contains no meaningful information related to the dialogue context or task.\\
    \midrule
    \textbf{\textsc{Step 2: Classify Statement Type}}\\
    \midrule
    \textbf{Factual Statement:} \\
    Makes claims about objective facts (e.g., dates, events, entities). Verifiable by directly comparing with the provided context (e.g., retrieved documents, dialogue history).\\
    Example: "stores accept various forms of unexpired identification, including ids from all us states."\\
    \midrule
    \textbf{Cognitive Statement:} \\
    Involves reasoning, interpretation, opinions, predictions, or subjective descriptions. Requires inference from context or indirect evidence.\\
    Example: "This practice ensures that they verify the age of their customers accurately and consistently"\\
    \midrule
    \textbf{\textsc{Step 3: Evaluate Factual Statements}}\\
    \midrule
    \textbf{Faithful:} Facts are supported by the context; no contradictions.\\
    \textbf{Invented:} Otherwise.\\
    \midrule
    \textbf{\textsc{Step 4: Evaluate Cognitive Statements}}\\
    \midrule
    Apply the following rules in sequential order: \\
    \textbf{Rule 1: Rational:} Whether the statement is plausible speculation.\\
    \textbf{Rule 2: Grounded:} Whether the statement is logically supported by the context or aligns with indirect evidence.\\
    \textbf{Rule 3: Unequivocal:} Whether the statement is the only reasonable conclusion supported by indisputable evidence, free from subjective bias.\\
    \midrule
    \end{tabular}
    \caption{Complete instructions for annotators identifying and evaluating statements in CogniBench.} 
    \label{fig:full_instruct}
\end{table*}

\subsubsection{Examples of Annotation}

\begin{figure*}[ht]
     \includegraphics[width=\textwidth]{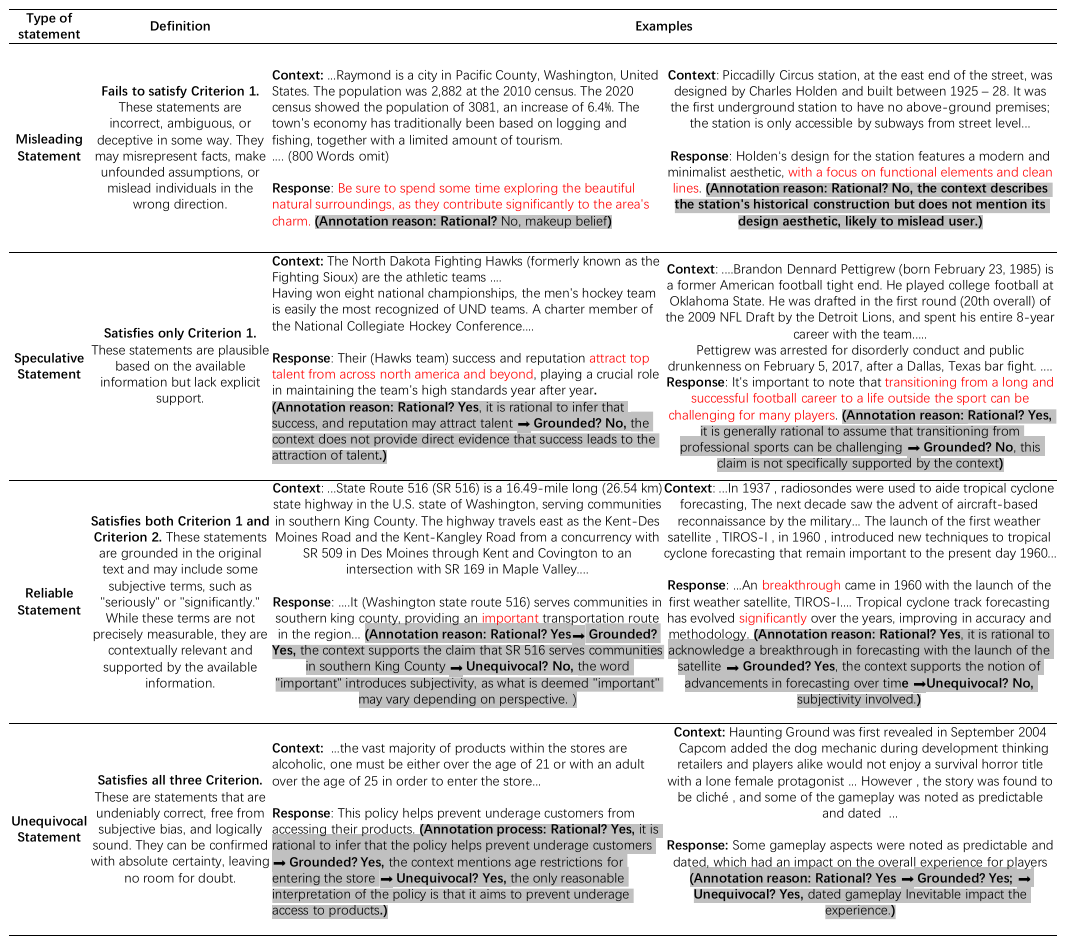}
     \caption{Examples of cognitive statements categorized according to the standards outlined in \cref{credibility:standards}. The red-highlighted text indicates issues with the statement, while the greyed-out text represents the corresponding annotation reasons.}
     \label{statment_rules}
\end{figure*}

We provide annotated examples in \cref{statment_rules} for four types of cognitive statements: Misleading, Speculative, Reliable, and Unequivocal. The context has been excerpted for ease of reading.

\subsection{Topic Distribution}

\begin{figure}[ht]
     \includegraphics[width=\linewidth]{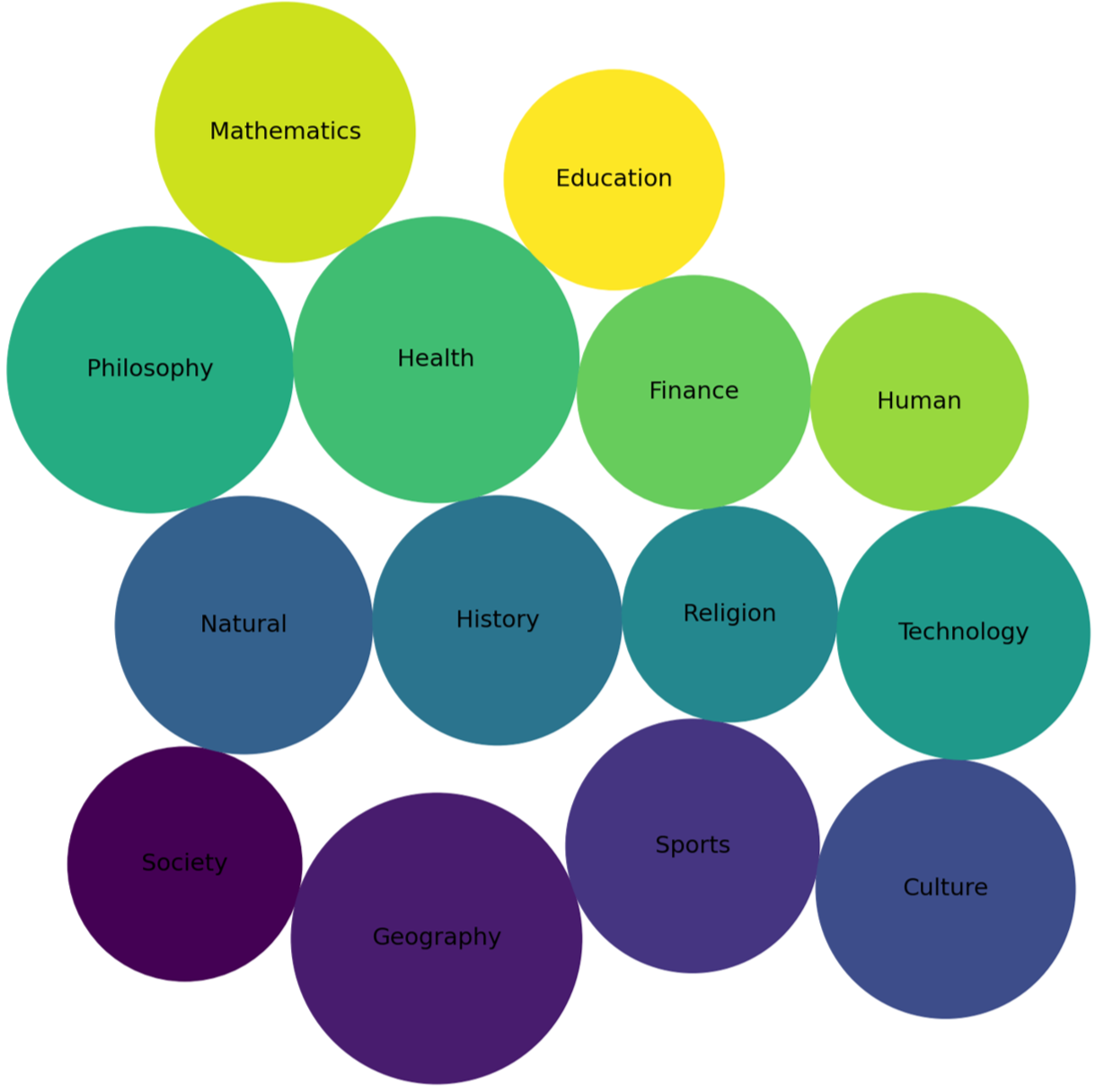}
     \caption{CogniBench employs a uniform sampling across various topics to ensure conversational diversity. The size of each bubble represents the proportion of that topic in the benchmark}
     \label{topic_distribution}
\end{figure}

As shown in \cref{topic_distribution}, to enhance the evaluation of large language models, we collect a diverse set of topics from Wikipedia, ensuring broad representation of real-world use cases in the knowledge-grounded multi-turn dialogues. This strategy allows us to rigorously test the generalization abilities of models evaluated on CogniBench.


\subsection{Prompt used to generate CogniBench-L}
\label{appendix:prompt}

We present the prompt used to generate CogniBench-L, as detailed in \cref{tab:system_user_prompts}. The user prompt is structured to first provide examples of both incorrect and correct annotations, covering both cognitive and factual statements. These examples are drawn from the Initial Prompting and Diagnosis process, where we identified recurring annotation patterns. The conversation is segmented into single-turn dialogues, with each sentence enclosed in < > and </> to facilitate sentence-level annotation. This structure allows the LLM to focus on annotating individual sentences. The sample prompt is then passed to GPT-4 five times, and a majority-vote approach is applied to determine the final annotation for each sentence.

\subsection{Implementation Detail}

\subsubsection{Hallucination Detection Methods Setup}
\label{app: baselines}

Using CogniBench, we conducted experiments with the following six distinct algorithms for hallucination detection:

\textbf{Tasksource} \cite{tasksource}: Tasksource is  zero-shot classification nli model, we pair each response sentence against context, and gather all "contradict" sentence as detection output.

\textbf{SelfCheckGPT} \cite{selfcheckgpt}: SelfCheckGPT implement use fine-tuned DeBERTa-v3-large, it output entailment (or Contradiction) score with input being the sentence and a sampled passage. We pair each response sentence against context, and gather all "contradict" sentence as detection output.

\textbf{FAVA} \cite{fava}:  FAVA (FAct Verification with Augmentation), a model for fine-grained hallucinations detections and editing. FAVA is trained on high-quality synthetic training data to identify hallucinations, incorporating retrieved knowledge. We map spans to sentence-level to ensure a fair comparison.

\textbf{RAG-Truth }\cite{niu2023ragtruth}: RAGTruth uses the Llama2 13B instruct fine-tuned with the training set from RAGTruth. The model takes the context-response pair with proper instructions as the input and treats the hallucinate span as the targeted generation output. We mapping spans to sentence-level to ensure a fair comparison.

\textbf{Auto-Labeling}: We employ  the auto-labeling pipeline \cref{sec:auto_labeling} on human-labeled CogniBench data to assess the reliability of this method.

\textbf{CogniDet}: We fine-tune the Llama3 8B instruct model \cite{llama3} using the CogniBench-L dataset, which is generated via the auto-labeling pipeline outlined in \cref{sec:auto_labeling}.

\subsubsection{CogniDet Training Details}

    We use the CogniBench-L that generated by auto-labeling pipeline proposed in \cref{sec:auto_labeling}, to fine-tune Llama3 8B instruct model \cite{llama3}. The training setup and hyper-parameters are as follows: the epoch is 3, the batch size is 2, the learning rate is 5e-5 We generate responses using sampling implemented via vLLM \cite{vllm}. Our model is trained on 8 NVIDIA A6000 GPUs. It takes approximately 18 hour to train.

    CogniDet takes the context-response pair and direct generate a list of hallucinated sentences as response, as shown in \cref{det_eg}.

\subsection{Evaluation of Auto-labeling}

\begin{table}[h!]
\resizebox{\columnwidth}{!}{
\begin{tabular}{@{}ccccccc@{}} 
\toprule
\textbf{\begin{tabular}[c]{@{}c@{}}Hallucination   \\ type\end{tabular}} &
  \multicolumn{2}{c}{\textbf{Overall}} &
  \multicolumn{2}{c}{\textbf{\begin{tabular}[c]{@{}c@{}}Factual   \\      Hallucination\end{tabular}}} &
  \multicolumn{2}{c}{\textbf{\begin{tabular}[c]{@{}c@{}}Cognitive   \\      Hallucination\end{tabular}}} \\ \midrule
\textbf{Method}                          & \textbf{Recall} & \textbf{Precision} & \textbf{Recall} & \textbf{Precision} & \textbf{Recall} & \textbf{Precision} \\ \midrule
\textbf{Auto-Labeling   (Threshold = 2)} & \textbf{77.98}  & 87.76              & \textbf{74.75}  & 91.05              & \textbf{78.56}  & 85.55              \\
\textbf{Auto-Labeling   (Threshold = 3)} & 75.88           & \textbf{89.63}     & 72.72           & \textbf{91.70}     & 76.43           & \textbf{87.83}     \\
\multicolumn{1}{r}{\textbf{- Sampling}}  & 67.72           & 88.05              & 67.98           & 89.50              & 66.76           & 86.33              \\
\multicolumn{1}{r}{\textbf{- CFP}}       & 60.49           & 85.11              & 53.69           & 85.26              & 62.65           & 84.29              \\ \bottomrule
\end{tabular}
}
\caption{Ablation Study of Auto-Labeling Pipeline}
\label{Tab:ablation}
\end{table}

We present an ablation study of our proposed auto-labeling pipeline in \cref{Tab:ablation}. We evaluate the impact of removing key components: (1) CFP refers to the removal of the Contrastive and Formative Prompting (CFP) approach as described in \cref{CFP}, and (2) Sampling indicates the exclusion of the Multi-response Sampling method introduced in \cref{Sampling}. The results highlight the importance of both techniques, as their absence leads to a noticeable drop in recall and precision, especially for cognitive hallucinations. For Auto-Labeling, we apply five-time sampling, where each sentence is annotated five times with identical prompts, and a majority-vote criterion is applied to determine its final classification. This offers flexibility in adjusting the trade-off between precision and recall.
The threshold refers to the decision boundary for classifying a sentence as faithful or hallucinated: a higher threshold increases precision by demanding stricter agreement across the responses, while a lower threshold enhances recall by relaxing the agreement requirement.

\subsection{Example of Hallucination Detection of CogniDet}
\label{appendix:CogniDet_example}

\begin{figure*}[h]
\centering
     \includegraphics[width=0.8\linewidth]{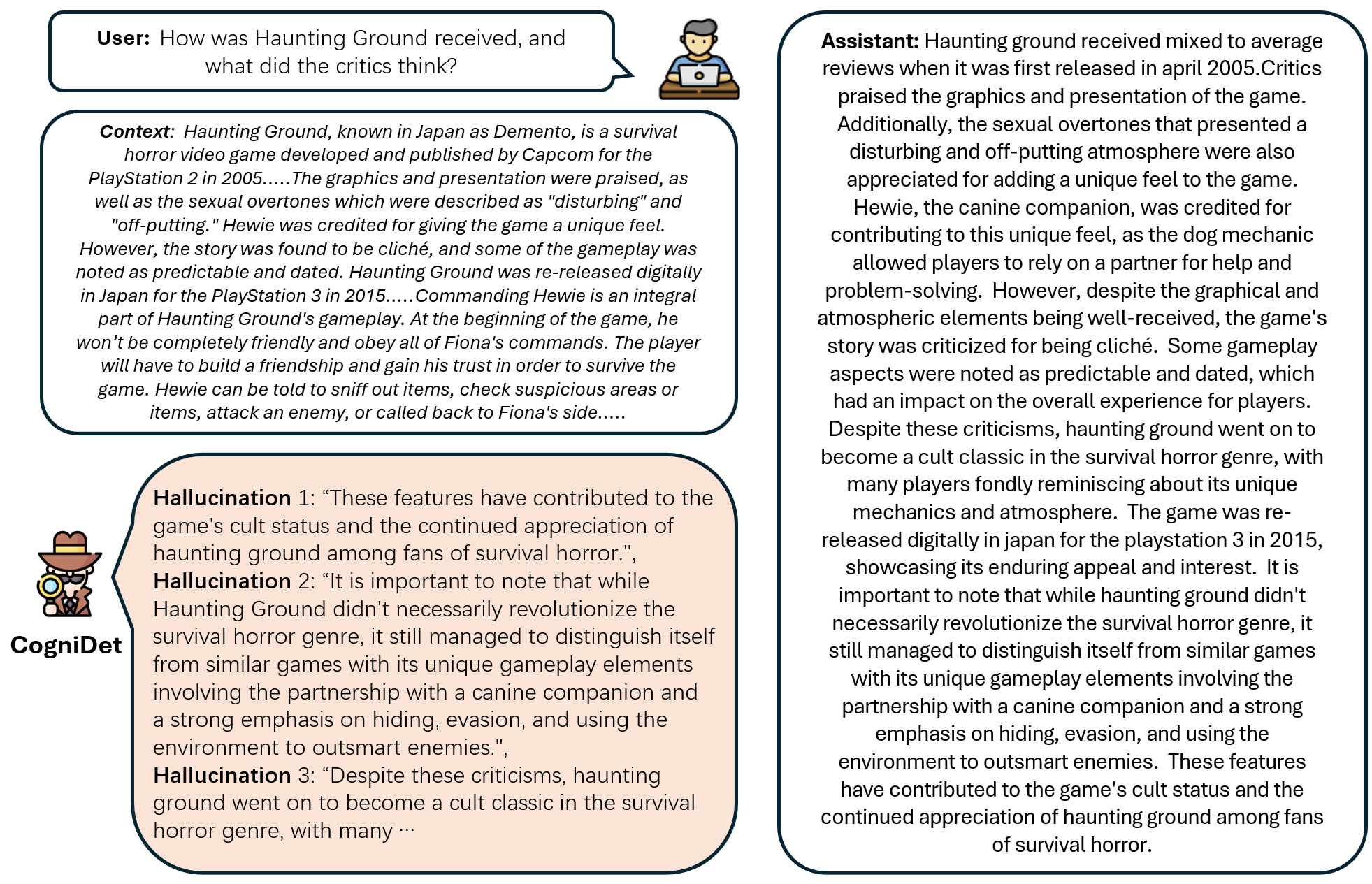}
     \caption{Example of hallucination detection output from CogniDet, trained on CogniBench-L}
     \label{det_eg}
\end{figure*}

We present a hallucination detection example in \cref{det_eg}. CogniDet takes Context and Dialogue as input and outputs a list of hallucinated statements, including invented, speculative, and misleading statements, all detected in a single forward pass.

\subsection{Performance of CogniDet versus training data size}

\begin{figure}[]
\centering
     \includegraphics[width=0.9\linewidth]{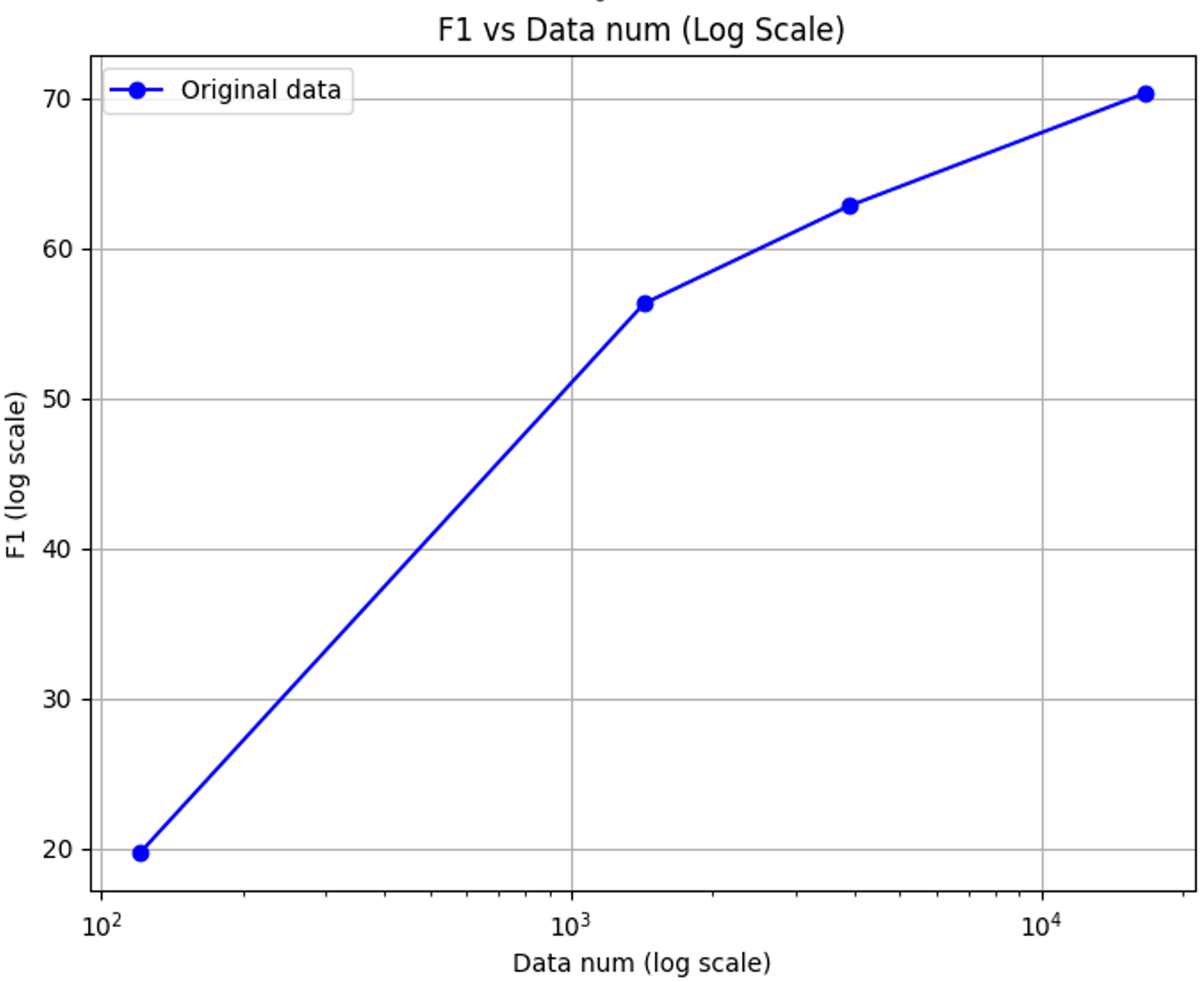}
     \caption{CogniDet's detection accuracy (F1) showing log-linear scaling with training data size}
     \label{fig:cogniDet_scaling}
\end{figure}

\cref{fig:cogniDet_scaling} reveals a clear log-linear relationship between training data size and model performance, suggesting continued performance improvements with additional data. This scaling law demonstrates the importance of large-scale annotated datasets like CogniBench-L for developing robust hallucination detection systems.

\subsection{Compare between CogniBench and existing benchmarks}

\begin{figure*}
     \includegraphics[width=\textwidth]{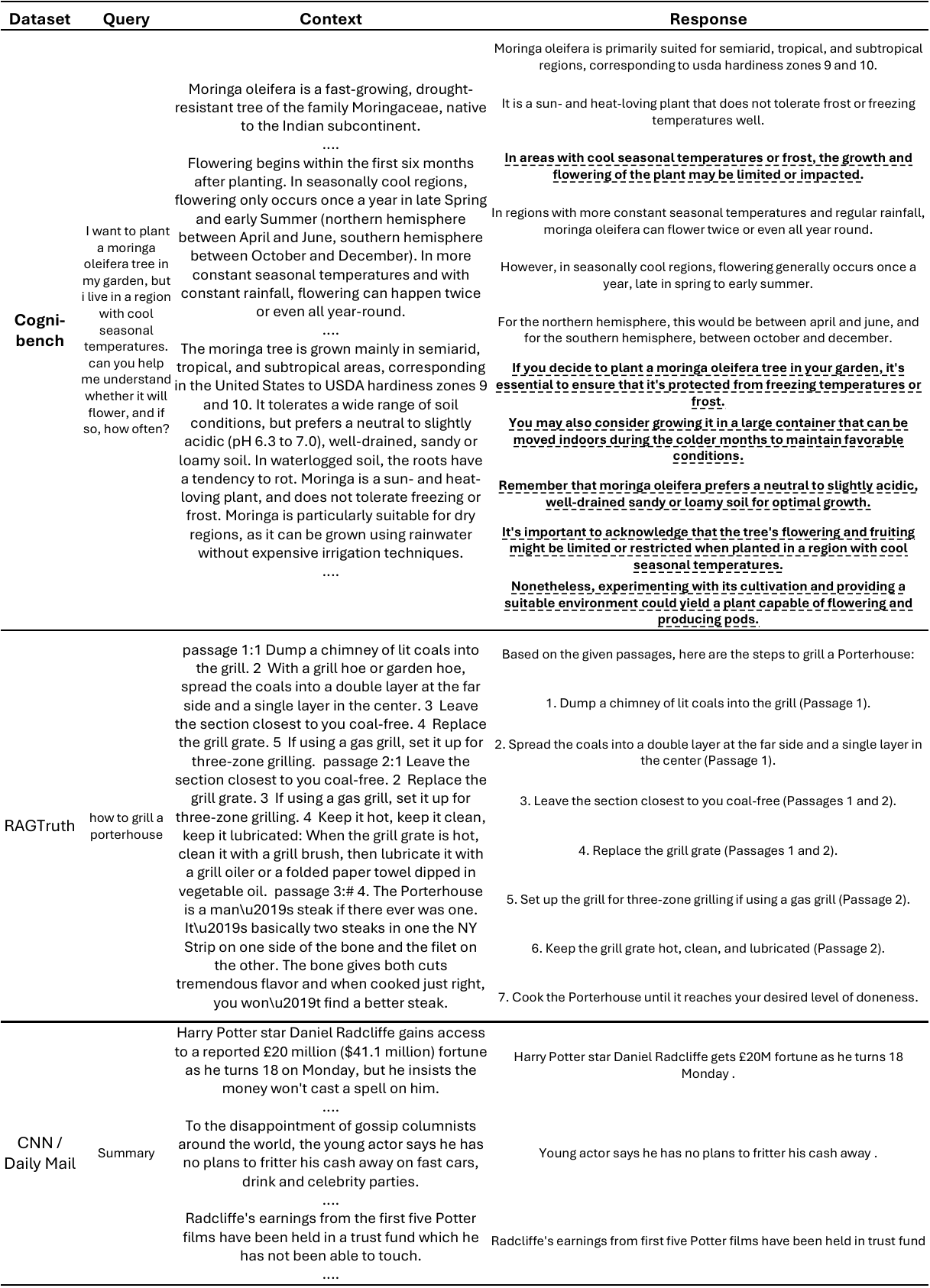}
    \caption{Example dialoge from CogniBench with RAGTruth \cite{niu2023ragtruth} and CNN/Daily Mail \cite{cnnmail}. Cognitive statements are bold and dotted underlined}
    \label{fig:dialogue_compare}
\end{figure*}

We compare the query context and response from RAGTruth \cite{niu2023ragtruth} and CNN/Daily Mail \cite{cnnmail} in \cref{fig:dialogue_compare},
Both RAGTruth and CNN/Daily Mail dataset are composed mainly with factual statement
RAGTruth answers focuses more on rephrase the information from the retrieved passages.
CNN/Daily Mail excels in news summarization, where the response retrieval key factual information form context to generate a condensing article.

CogniBench differentiates itself from these previous datasets by not only emphasizing factual recall but also generating cognitive statements—such as inferences, evaluations, and opinions. For example, in the case of Moringa oleifera cultivation (shown in \cref{fig:dialogue_compare}), the generated response includes not only factual details about its growing conditions but also reasoning about how climate and soil factors can influence flowering patterns. This type of response moves beyond simple fact recall, offering more complex insights that align with cognitive processes like evaluation and inference.

\begin{table*}[!ht]
\renewcommand{\arraystretch}{1.1}
\small
\centering
    \begin{tabular}{p{1.0\textwidth}}
    \toprule
    \textbf{\textsc{System Prompt}}\\
    \midrule
    Your task is to evaluate the faithfulness of claims. Tag the statements enclosed in the empty tag spaces provided by the user.\\
    \textbf{Tags to Use:}\\
    \textbf{<faithful>:} The sentence is fully verified by the references of the conversation.\\
    \textbf{<hallu>:} hallucination, use this tag for statements that reflect unreliable personal opinions or judgments or statements that include information not found in the references or context\\
    \textbf{<irrelevant>:} The sentence is irrelevant to the reference or does not provide any useful information.\\
    \midrule
    \textbf{\textsc{User Prompt}}\\
    \midrule
    \textbf{Example of incorrect annotation:} \\ 
    \textbf{Reference:} \\
    Jacques Michel Gabriel Paul Benoist-Méchin (1 July 1901 – 24 February 1983) was a French far right politician and writer...
    \textbf{Claims:} \\
    This claim should be tagged as hallu: "In 1936, Jacques benoist-méchin joined the french popular party..." since french popular party is far-right political group opposed to democracy is not mentioned in the reference even though Jacques benoist-méchin was indeed a far right politician. \\

    \\

    \textbf{Example of incorrect annotation:} \\
    \textbf{Reference:} The title-safe area or graphics-safe area is, in television broadcasting, a rectangular area which is far enough in from the four edges, such that text or graphics show neatly... \\
    \textbf{Claims:} \\
    This claim should be tagged as faithful: "Content placed in Title-safe area remains legible and undistorted..." since the claim can be reasonably inferred from the reference. \\
    \\
    \textbf{You SHOULD refer to the following correct examples:}  \\
    \textbf{Reference:} \\
    Izumi Station is a railway station on the Jōban Line in the city of Iwaki, Fukushima, Japan, operated by East Japan Railway Company (JR East). The station also has a freight depot for the Fukushima Rinkai Railway Main Line.\\
    \textbf{Claims:} \\
    <>Izumi station has a combination of one island platform and one side platform, which are connected to the station building by a footbridge.</> <>The island platform allows trains to pass on either side, while the side platform is used for trains arriving from just one direction.</> <>This layout ensures efficient handling of the passenger and freight trains that pass through the station.</>\\
    \textbf{Annotation:} \\
    <faithful>Izumi station has a combination of one island platform and one side platform, which are connected to the station building by a footbridge.</faithful> <hallu>The island platform allows trains to pass on either side, while the side platform is used for trains arriving from just one direction.</hallu> <hallu>This layout ensures efficient handling of the passenger and freight trains that pass through the station.</hallu>\\
    \\
    Now, based on the examples above, please evaluate the following claim: \\
    \textbf{Reference:}
    \{reference\} \\
    \textbf{Claims}: 
    \{dialogue\} \\
    \bottomrule
    \end{tabular}
    \caption{System and user prompts for assessing the faithfulness of claims.}
    \label{tab:system_user_prompts}
\end{table*}

\end{document}